%% file: colm2025_conference.tex
\documentclass{article} % For LaTeX2e
\usepackage{url}
\usepackage[final]{colm2025_conference}\usepackage{microtype}

\usepackage{booktabs}
\usepackage{graphicx}

\usepackage{lineno}

\usepackage[utf8]{inputenc} % allow utf-8 input
\usepackage[T1]{fontenc}    % use 8-bit T1 fonts
\usepackage{booktabs}       % professional-quality tables
\usepackage{amsfonts}       % blackboard math symbols
\usepackage{nicefrac}       % compact symbols for 1/2, etc.
\usepackage{subfigure}
\usepackage{diagbox}
\usepackage{multirow}
\usepackage{caption}
\usepackage{xcolor}
\usepackage{colortbl}
\usepackage[graphicx]{realboxes}

\usepackage{hyperref}

\definecolor{darkblue}{rgb}{0, 0, 0.5}
\hypersetup{colorlinks=true, citecolor=darkblue, linkcolor=darkblue, urlcolor=darkblue}

\title{\raisebox{-0.1\height}{\includegraphics[height=1.0em]{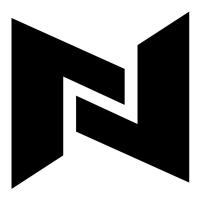}}
Nex-N1:  Agentic Models Trained via a Unified Ecosystem for Large-Scale Environment Construction}

\author{Nex-AGI Team}

\begin{document}

% \ifcolmsubmission
% \linenumbers
% \fi

\maketitle

\input{chapters/abstract}
\input{chapters/intro}
\input{chapters/method}
\input{chapters/results}

\bibliography{colm2025_conference}
\bibliographystyle{colm2025_conference}

\input{chapters/appendix}

\end{document}

%% file: chapters/abstract.tex
\begin{abstract}
The evolution of Large Language Models (LLMs) from passive responders to autonomous agents necessitates a fundamental shift in learning paradigms—from static imitation to incentive-driven decision making. However, this transition is significantly impeded by the lack of scalable infrastructure capable of constructing high-quality interaction signals for effective policy learning.
To address this, we introduce a comprehensive method designed to systematically scale the diversity and complexity of interactive environments.
Our method realizes this scaling by addressing three orthogonal dimensions: (1) Complexity: NexAU, a flexible agent framework that supports building complex agent hierarchies via simple configurations; (2) Diversity: NexA4A automatically generate diverse agent hierarchies from natural language to cover infinite domains; and (3) Fidelity: NexGAP bridges the simulation–reality gap by integrating dynamic real-world environment for grounded trajectories synthesis.
We train Nex-N1 upon the diverse and complex interactive environments established by our infrastructure.
Empirical results on benchmarks such as SWE-bench and $\tau^2$ demonstrate that Nex-N1 consistently outperforms SOTA open-source models and achieves competitive performance frontier proprietary models on complex agentic tasks. We open-source the Nex ecosystem and model weights to facilitate further research\footnote{https://github.com/nex-agi/Nex-N1}.
\end{abstract}

\begin{figure}[htbp]
    \centering
    \includegraphics[width=1.0\linewidth]{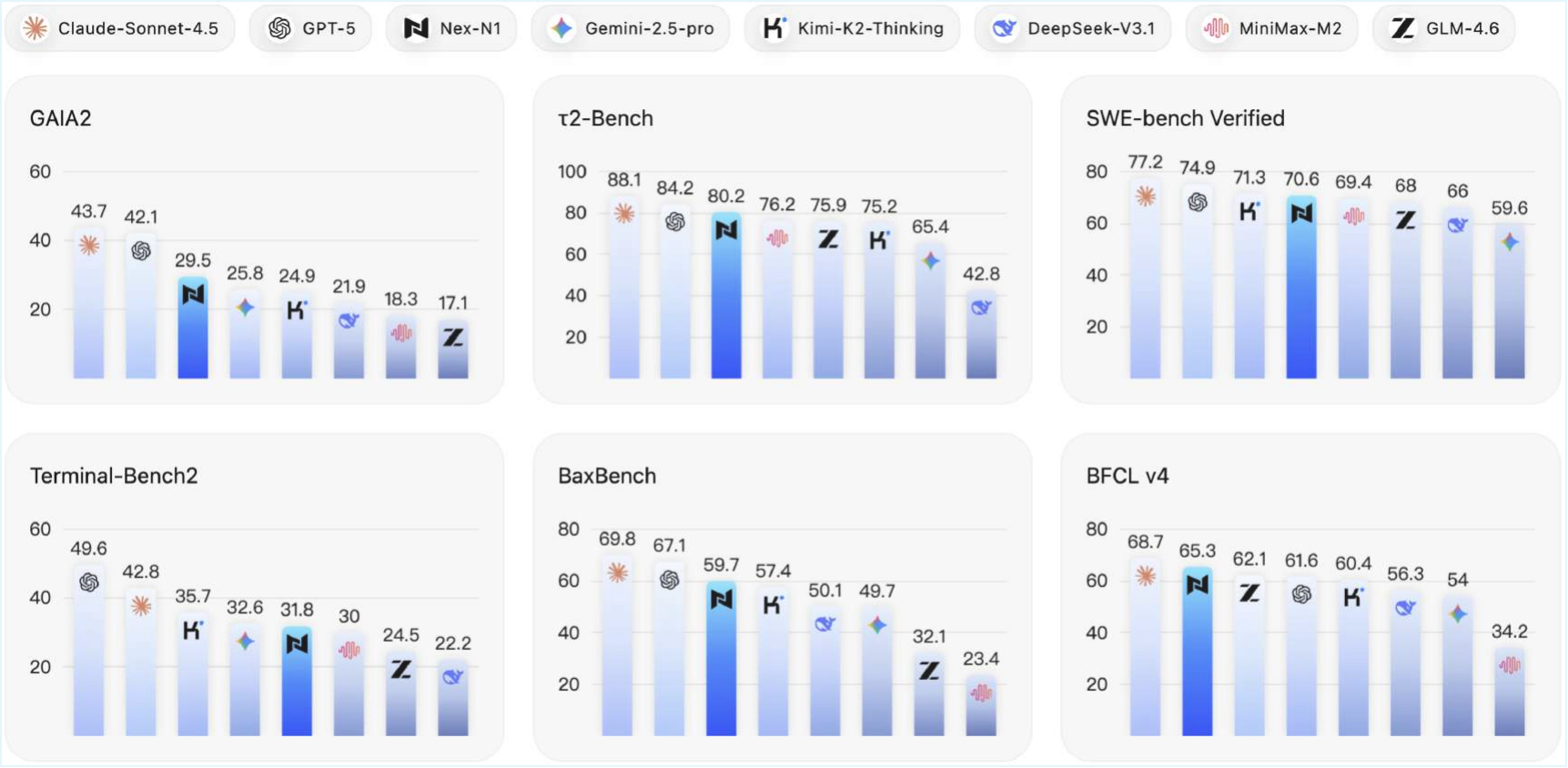}
    \caption{Performance comparison of Nex-N1 against state-of-the-art models on agent and coding benchmarks.}
    \label{fig:placeholder}
\end{figure}

%% file: chapters/intro.tex
\section{Introduction}

The evolution of Large Language Models (LLMs) from passive information processors to autonomous Agents represents a fundamental shift in the pursuit of Artificial General Intelligence (AGI) \citep{xi2023rise, wang2023voyager, bubeck2023sparks}. While current foundation models demonstrate remarkable capabilities in knowledge representation and reasoning \citep{openai2023gpt4, liu2024deepseek}, deploying them as reliable agents in real-world scenarios remains a formidable challenge. 
A central capability required for this transition is 'Agency'—the capacity to anchor deep reasoning in reality: moving beyond internal intuition to actively perceive the environment and adapt strategies based on dynamic feedback.

% A central capability required for this transition is ``Agency''—the capacity to perceive dynamic environments, reason over long horizons, and execute multi-step plans.

However, a critical misalignment exists between the myopic ``next-token prediction'' objective governing LLM pre-training and the long-horizon, goal-oriented nature of agentic tasks. 
We argue that bridging this gap requires transforming the learning process itself: from learning what to say to learning how to act, which demands a new scale of interactive environments.

\begin{itemize}
   \item \textbf{Scarcity of Diverse Environments.} LLMs trained on static text corpora often act as ``System 1'' responders, lacking the ``System 2'' rigor required for complex planning \citep{bengio2021system2}. Without exposure to environments that demand long-term reasoning, models succumb to probability traps and myopic decision-making \citep{valmeekam2023planning}. Furthermore, constructing interactive environments that are both broad in scope and reliable in structure is prohibitively expensive. Current approaches rely on limited environments or rigid frameworks \citep{liu2023agentbench, andrews2025are}, which fail to provide the behavioral diversity needed for models to generalize to novel tasks \citep{barres2025tau2}.
    
   \item \textbf{Lack of Realistic Grounding.} Agents trained on purely synthetic or static data often struggle with the complexity of real-world execution. They exhibit a disconnect between ``thought'' and ``action'', leading to hallucinations in tool usage—such as invoking APIs based on outdated assumptions \citep{patil2023gorilla, schick2023toolformer}. Unlike biological systems that learn through interaction, LLMs typically fail to perform robust error recovery or self-correction when actions fail \citep{shinn2023reflexion, yao2023react}. True agentic capability requires training on trajectories that capture the latency, stochasticity, and feedback loops of real-world execution.
\end{itemize}

To address this, we propose an approach based on \textbf{agentic scaling} to generate diverse environments and high-quality data. We introduce a unified system comprising three components: NexAU (Agent Universe), a universal agent framework that hides complexities of agent features (execution loop, tools, sub-agents, context management etc.) from agent builders. Simple configurations could generate very complex and diverse agents. NexA4A (Agent for Agent), a generative system that automatically synthesizes diverse agent architectures and workflows from natural language specifications; and NexGAP (General Agent-data Pipeline), which leverages real-world Model Context Protocol (MCP) tools and information fusion to generate massive-scale, end-to-end trajectories rooted in authentic execution.
Leveraging this system, we train \textbf{Nex-N1}, a series of models that demonstrate robust generalization across heterogeneous agent frameworks. Our main contributions are summarized as follows:
\begin{itemize} 
\item \textbf{Infrastructure for Environmental Scaling.}
We propose a unified ecosystem (NexAU, NexA4A, NexGAP) that transforms environment construction from manual engineering to automated synthesis. By treating agent environments as generative language specifications rather than static code, we break the dependency on human-designed enviroments and enable the infinite scaling of diverse interaction topologies.

\item \textbf{State-of-the-Art Agentic Performance.} 
% We introduce Nex-N1, an agentic language model trained on the synthesized data. 
Extensive evaluations on benchmarks such as SWE-bench \citep{jimenez2024swebench}, GAIA 2 \citep{andrews2025are}, and BFCL \citep{patil2025bfcl} show that Nex-N1 significantly outperforms open-source models of comparable size and achieves competitive performance against SOTA open-source models in agentic capabilities.
    
\item \textbf{Robustness and Practicality.} 
% We demonstrate that Nex-N1 is not overfitted to a specific framework. 
Nex-N1 shows strong generalization across different agent frameworks (e.g., OpenHands \citep{wang2024openhands}, Claude Code) and excels in practical human evaluations for tasks like end-to-end webpage creation and autonomous research.
    
\item \textbf{Open-sourcing.} To accelerate community research in agentic scaling, we open-source the Nex-N1 model weights, the inference code, and a subset of our high-quality agentic training data.
\end{itemize}

%% file: chapters/method.tex
\section{Method}
To advance the model’s agentic capabilities including its understanding of environments, effective tool use, and long-term planning that ultimately enables it to complete complex real-world tasks, we pursue an approach centered on \emph{agentic scaling}.

\begin{figure}[htbp]
    \centering
\includegraphics[width=0.98\linewidth]{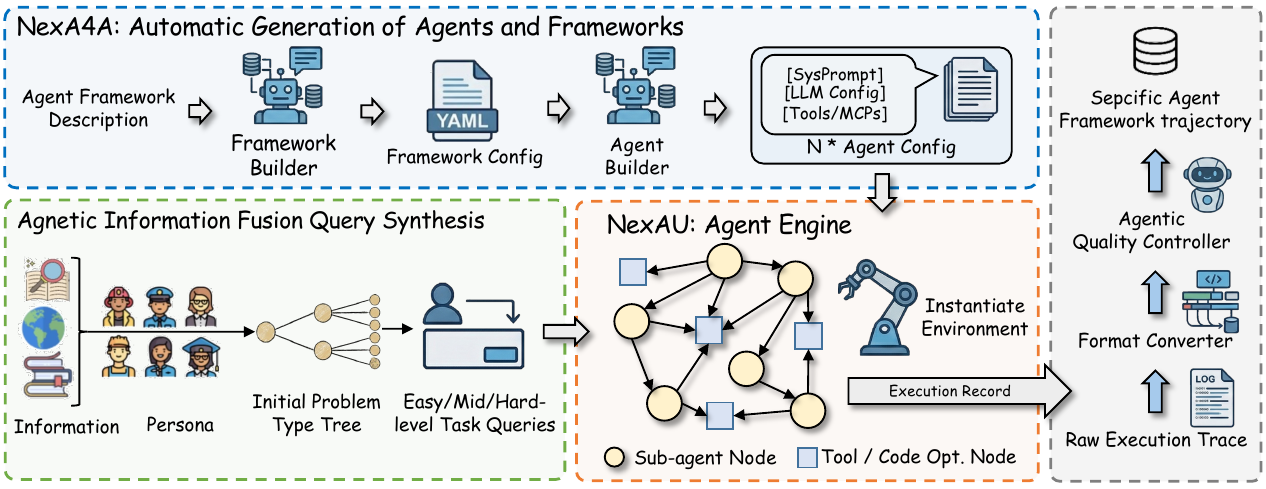}
    \caption{NexA4A agent framework workflow: a comprehensive architecture diagram from framework description to high-quality trajectory generation.}
    \label{fig:placeholder}
\end{figure}

\subsection{NexAU: A Modular Runtime for Scalable Agent Framework}
\label{sec:nexau}
% Scaling agentic capabilities requires environments that are both broad in scope and reliable in structure—conditions under which models can learn to plan, act, and coordinate in ways that generalize beyond handcrafted demonstrations. A natural path is to build upon the growing collection of open-source agent frameworks. However, when examined at scale, this ecosystem exposes two structural obstacles:
Agent capabilities emerge from interaction. An agent learns to perceive, plan, and act only by engaging with environments that provide structure, feedback, and opportunities for tool use. Therefore, scaling these capabilities requires environments that are both broad in scope and consistent in structure. A natural path is to build upon the collection of open-source agent frameworks. However, when examined at scale, this ecosystem exposes two structural obstacles:
\textbf{a)} \textit{the lack of scalable and stable environment simulation}, including tools, state transitions, and error behaviors, requires substantial manual effort, as most frameworks were designed for small-scale experiments rather than large-scale, reproducible trajectory generation;
\textbf{b)} \textit{existing frameworks provide limited diversity}. They cover only a narrow range of tasks, tools, and interaction patterns, with variable implementation quality, and the resulting behavioral space for agentic learning is constrained.

These constraints prevent existing environment resources from supporting the robustness and coverage required for large-scale agentic learning. Scaling agent capabilities ultimately demands environments that can automatically scale, remain stable under large-scale execution, and offer high diversity—capabilities that current frameworks fail to provide. Yet despite they differ substantially in design, the underlying competency a model must learn is remarkably consistent: conditioned on a context, determine the next action. Many implementation-level distinctions (e.g., whether context is transmitted through function-call arguments or shared state) are inconsequential. What truly matters are the context-passing relationships among agents, each agent’s system prompt, the tools involved, and the invocation format.

This observation motivates a unified and extensible agent engine that abstracts away framework-specific idiosyncrasies while preserving the behavioral structure essential for learning. So we introduce NexAU (Nex Agent Universe), a lightweight, high-throughput runtime that decouples agent definition from agent execution. NexAU provides a consistent substrate for faithfully simulating tools, environments, and error dynamics at scale, enabling the generation of high-quality trajectories across diverse tasks. Unlike rigid graph-based orchestration systems, NexAU adopts a recursive, fractal architecture inspired by the ReAct paradigm \cite{yao2023react}, treating sub-agents, tools, and external services as interchangeable functional units. This design both unifies heterogeneous frameworks under a single execution model and expands the diversity and realism of environments available for agentic scaling.

\paragraph{Recursive agentic loop and hierarchical decomposition.}
At the core of NexAU is the standardized agentic execution loop. Rather than enforcing a static computational graph, NexAU is able to model complex behavior through dynamic, hierarchical recursion. \textbf{a)}\textit{The Agentic Loop.} Each agent instance operates a localized ReAct loop, where it perceives context, reasons via the LLM, and executes an action. \textbf{b)}\textit{Recursive Delegation.} A distinct feature of NexAU is unifying tools and sub-agents. To the parent agent, a sub-agent is simply a tool with a defined input schema. When a parent invokes a sub-agent, the runtime instantiates a child execution context with its own system prompt, state, and distinct toolset. This child runs its own ReAct loop until a termination condition is met, returning the result to the parent's observation stream. \textbf{c)}\textit{Isolation and State}: This recursive structure ensures that reasoning states are isolated. A sub-agent's "Thought" trace does not pollute the parent's context window, allowing NexAU to simulate extremely long-horizon tasks without context overflow. (e.g., a "CTO" agent delegating to a "Software Engineer" agent) 

\paragraph{Decoupled composition via declarative schema.}
To achieve the diversity required for general agentic learning, NexAU separates the \textit{logical topology} of an agent system from its \textit{imperative implementation}. Agents are defined via declarative YAML configurations, while the runtime handles the binding of logic. This decoupling is critical for \textbf{NexA4A} (our generative environment system, to be described in Section \ref{sec:nexa4a}), as it allows the programmatic synthesis of new agent architectures without generating executable code. A unified schema defines the agent's persona, capabilities, and hierarchical relationships:
\begin{figure}[t]
    % \centering
\includegraphics[width=1\linewidth]{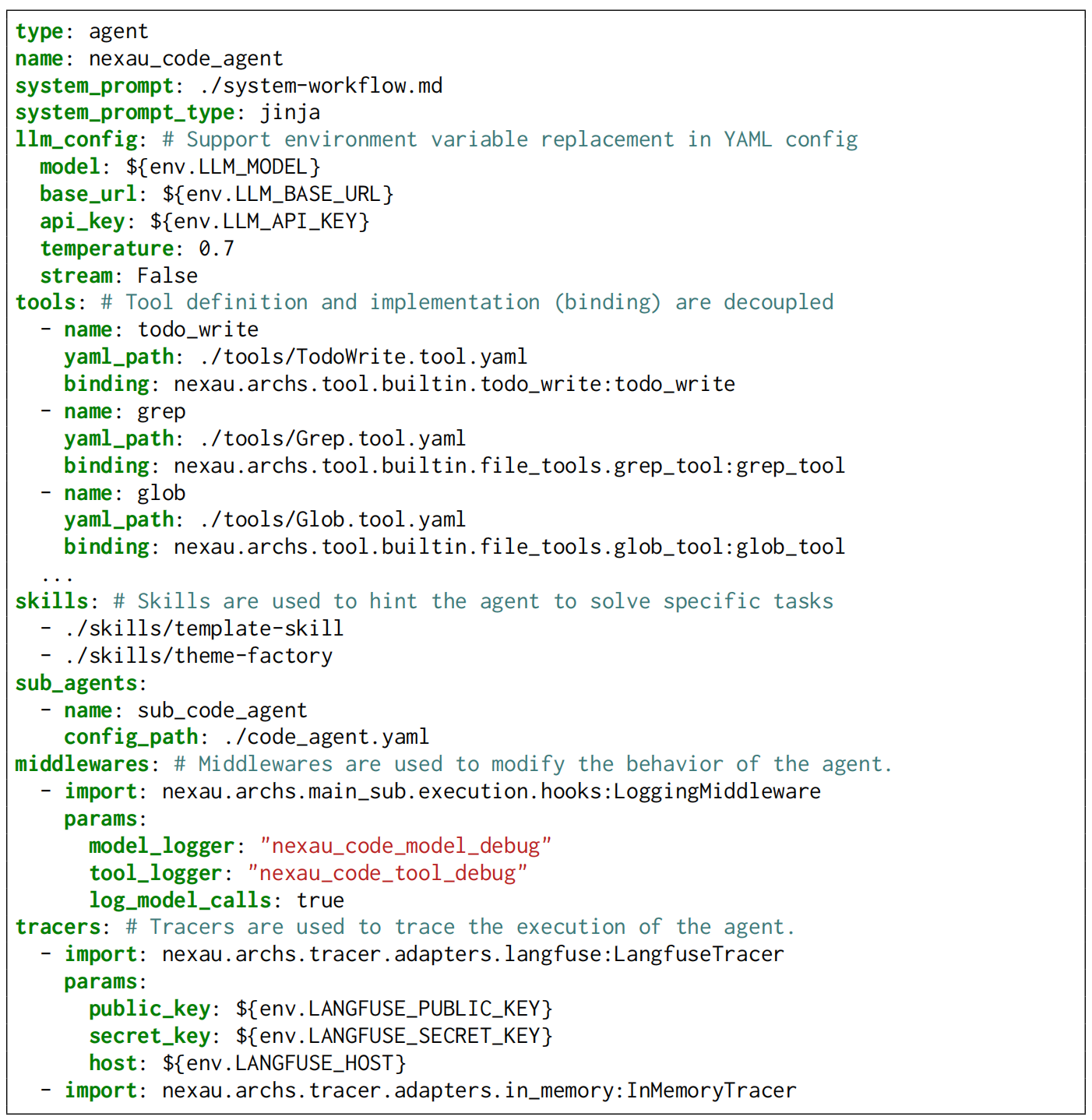}
\caption{A simplified NexAU definition. Sub-agents are composed alongside standard tools, enabling arbitrary depth of composition.}
    \label{fig:nexau_yaml}
\end{figure}

By manipulating these configurations, we can procedurally generate environments ranging from single-agent tool loops to multi-agent organizations, creating a diverse distribution of "interaction topologies" for the model to learn.

\paragraph{Unified interface for diverse capabilities.}
NexAU abstracts various interaction mechanisms into a unified interface to maximize the model's exposure to different environments during training.  \textbf{a)}\textit{MCP Support.} To ground agents in real-world data, NexAU integrates MCP to connect agents to live external servers (e.g., GitHub, Database interfaces) via a standardized protocol. This ensures that the data synthesis pipeline (NexGAP) respects the latency, error modes, and statefulness of real-world APIs. \textbf{b)}\textit{Skills.} NexAU supports \textit{Skills}—self-contained directories containing specialized prompts, few-shot examples, and executable scripts (similar to Claude Skills~\cite{claude_skills}). These are injected into the agent's context dynamically, teaching the model to utilize "retrieved" procedural knowledge. \textbf{c)}\textit{Global Storage and Memory.} While the ReAct loop is stateless between steps, NexAU provides a thread-safe \texttt{GlobalStorage} mechanism. This allows agents and tools to persist state across the recursion tree, enabling the simulation of complex state transitions (e.g., a file system changing state across multiple sub-agent edits).

By standardizing these diverse features into a coherent execution stream, NexAU transforms the chaotic landscape of agent frameworks into a structured data generator, providing the high-fidelity trajectories necessary for training Nex-N1.

\subsection{NexA4A: Automatic Generation of Agents and Frameworks}
\label{sec:nexa4a}
Scaling agentic capabilities requires not only high-quality environments but also a broad distribution of agent behaviors and workflow structures. To expand this space systematically, we introduce NexA4A (Agent for Agent), a system that generates agents as well as the entire agent frameworks from natural-language specifications. Each generated agent is defined by a system prompt outlining its role and meta-skills, where each meta-skill corresponds to a set of actions implemented through NexAU components such as sub-agents, tools, or MCPs. NexA4A automates this construction process: it generates system prompts, creates sub-agents, selects or synthesizes tools, integrates MCPs, and produces a complete agent configuration via the AgentBuilder.

Beyond individual agents, NexA4A can automatically generate full-blown multi-agent frameworks. It constructs a declarative configuration that defines nodes and their interaction structure using a component called FrameworkBuilder. A central MetaAgent interprets a high-level description of the target framework, chooses an appropriate workflow pattern, decomposes it into hierarchical tasks, and assembles multi-agent systems—typically one to three layers deep. By coordinating the FrameworkBuilder, AgentBuilder, and supporting sub-agents, NexA4A designs workflows, assigns roles, allocates meta-skills, decomposes skills into executable actions, configures agents, and synthesizes any required custom tools.

This pipeline enables the large-scale creation of diverse agents and, crucially, diverse agent frameworks, providing the breadth and variability needed for robust agentic scaling.

\subsection{NexGAP: End-to-End Agentic Data Generation}
With scalable environments established, we introduce NexGAP (General Agent-data Pipeline) to generate end-to-end agentic trajectories that expose models to the full range of behaviors needed for complex task solving: hierarchical decomposition, planning, tool-enabled problem solving, and iterative refinement based on environment feedback.
NexGAP begins by leveraging real MCP tools as capability seeds to produce framework-construction queries for NexA4A. After NexA4A synthesizes the agent framework, we apply Information Fusion Query Synthesis to generate tasks of varying difficulty tailored to that framework. NexAU then executes the resulting agents within their environments, producing raw interaction traces. To maintain interoperability across heterogeneous designs, NexGAP normalizes trajectories into multiple tool-call formats (including the OpenAI format and several XML-based variants).
This pipeline enables large-scale generation of realistic agentic data. In total, we construct over 200 agent frameworks and environments, whose agent and sub-agent graphs range from 1 to 34 nodes, spanning standard ReAct agents, multi-layer multi-agent systems, and fixed workflow pipelines. The resulting trajectories cover seven distinct tool-call formats, providing broad coverage of interaction patterns and execution semantics. This diversity forms a robust foundation for training models to operate reliably across a wide spectrum of agentic settings.

\paragraph{Incorporating real MCP tools for authentic environments.}
Custom tools generated by NexA4A are often simple code snippets which have limited interaction with outer systems such as APIs and databases. To expose agents to realistic system-level environments, we incorporate real, callable MCP tools. We collect MCP tools from public repositories, perform extensive cleaning and validation, and ultimately curate over one hundred high-quality, production-ready tools. Using large-scale real-world agent scenarios gathered from the web, we extract, cluster, and refine them into hundreds of high-fidelity interaction patterns. These patterns are used to construct scenario-aligned framework queries, paired with the curated MCP tools, enabling fully authentic, end-to-end agent frameworks that reflect real-world interaction workflows.

\paragraph{Information fusion for query synthesis.}
\label{sec:query}
We develop a query-synthesis framework capable of generating diverse, high-quality queries representative of real-world scenarios. The framework uses a \emph{Problem Type Tree}—a hierarchical, bilingually annotated taxonomy—to provide structured problem categories and ensure comprehensive coverage of the query space. To mitigate sampling bias, we adopt an inverse-frequency weighted strategy that increases sampling probabilities for underrepresented categories.
To diversify query generation, each query is conditioned on four variables: user persona, problem type (sampled from the taxonomy), framework context, and difficulty level (easy/medium/hard). The framework uses a multi-agent orchestration design composed of: (1) a rewriting agent that ensures persona–problem-type alignment, (2) a query synthesis agent that generates stratified queries by difficulty, and (3) optional enhancement agents that provide web-grounded context or apply fuzzification to increase naturalness. The taxonomy can be dynamically expanded through LLM-guided generation of new problem types, constrained by parent-node semantics and framework capabilities. This systematic approach yields scalable, realistic, and semantically rich queries reflective of real-world user interactions.

\subsection{Agentifying Non-Agent Data Construction}
For certain non-agent tasks of interest, we further leverage agentic processes in query construction, response generation, and data quality checking to produce higher-quality datasets and train more capable models.

\paragraph{Search-enhanced agentic data construction.}
Grounding query synthesis in retrieved knowledge is essential for two reasons: large language models have temporal and domain limitations in their training data, and unguided generation is prone to hallucinations. For example, producing physically implausible scenes in educational simulations. By integrating web search augmentation into the synthesis pipeline, we establish a factual foundation that anchors generation in verifiable information. Search grounding reduces knowledge errors, enhances scenario completeness, and ultimately produces more realistic and context-rich queries. This evidence-based paradigm yields higher-quality synthetic data and provides a more reliable benchmark for agent evaluation.

\paragraph{Supervisor tool feedback and quality optimization.}
We implement a supervisor as a tool that dynamically updates its understanding of the environment and refines subsequent planning. This design addresses two challenges: (1) initial code generated for complex scenarios is often unusable, requiring self-repair mechanisms; and (2) rendered outputs frequently exhibit visual issues, especially in complex 3D or interactive scenes. To resolve this, we use multimodal models for visual feedback.

Despite significant improvements, visual feedback occasionally proved unreliable and code repair sometimes failed. To mitigate these issues, we introduced engineering optimizations: converting visual feedback from continuous scoring to binary judgments (e.g., whether a scene is too dark or whether a page is complete), thus turning subjective aesthetics into objective criteria; and imposing a maximum repair-iteration limit, discarding code that cannot be fixed. Notably, the supervisor need not rely solely on vision-language models—agent-based architectures, GUI agents, or RL-enhanced feedback loops offer promising alternatives.

\paragraph{Trajectory quality assessment.}
To ensure the quality of training data, we filter trajectories exhibiting truncation, repetition, hallucination, or reward hacking. Agentic trajectories are significantly more challenging than standard post-training data due to their length and scenario diversity, motivating the use of Quality Assessment Agent for scalable assessment. We manually inspect trajectories across settings to construct a taxonomy focused on instruction following, failure analysis, trace anomalies, and tool-design issues. The agent is prompted with this taxonomy, and outputs identified issues in JSON format.

To remain within context length and improve accuracy, the judge processes trajectories iteratively, receiving only a small batch of messages and its previous assessment at each step. Assessment over large samples reveals issues such as ineffective tool design (e.g., placeholder outputs), excessive tool-return verbosity, and widespread reward hacking in coding agents (e.g., fabricating test results using nonexistent test files). These insights guide improvements to the agent framework and reduce undesirable behaviors in the final trained model.

%% file: chapters/results.tex
\section{Results}

\subsection{Benchmark Evaluation}

We select six representative benchmarks across the dimensions of general and specialized capabilities to comprehensively evaluate the performance of the Nex-N1 model in agentic tasks. The results in Table~\ref{tab:main result} Nex-N1 series consistently outperforms other open-source models of comparable size, with the largest model surpassing GPT-5 in tool use.

\begin{table}[htbp]
    \centering
    \caption{Agentic benchmark results and comparative analysis.} 
    \label{tab:main result}
    
    \resizebox{\textwidth}{!}{%
        \begin{tabular}{l c c c c c c} 
            \toprule
            \textbf{Model} & \textbf{$\tau^2$} & \textbf{GAIA2} & \textbf{SWE-bench} & \textbf{Terminal Bench 2} & \textbf{Baxbench} & \textbf{BFCL v4} \\
            \midrule
            \rowcolor{lightgray!20}\multicolumn{7}{c}{\emph{Proprietary Models}} \\ 
            Claude-Sonnet-4.5 &88.1 &43.7 & 77.2 & 42.8 & 69.8 & 68.8 \\
            GPT-5 &84.2 &42.1 & 74.9 & 49.6 & 67.1 & 61.6 \\
            Gemini-2.5-pro &65.4 &25.8 & 59.6 & 32.6 & 49.7 & 54.0 \\
            \midrule
            \rowcolor{lightgray!20}\multicolumn{7}{c}{\emph{Open Source Models > 100B}} \\ 
            GLM-4.6 &75.9 &17.1 & 68.0 & 24.5 & 32.1 & 62.1 \\
            Minimax-M2 &76.2 &18.3 & 69.4 & 30.0 & 23.4 & 34.2 \\
            Kimi-K2-thinking &75.2 &24.9 & 71.3 & 35.7 & 57.4 & 60.4 \\
            DeepSeek-V3.1 &42.8 &21.9 & 66.0 & 22.2 & 50.1 & 56.3 \\
            DeepSeek-V3.1-Nex-N1 &80.2 &29.5 & 70.6 & 31.8 & 59.7 & \textbf{65.3} \\
            \midrule
  
           \rowcolor{lightgray!20}\multicolumn{7}{c}{\emph{Open Source Models < 100B}} \\
            Qwen3-32B & 41.5 & 10.8& 12.9 & 7.9 & 35.6 & 45.4 \\
            Qwen3-32B-Nex-N1 &72.1  &16.7 &50.5	&16.7 & 34.8 & \textbf{60.5} \\
            Qwen3-30B-A3B & 28.5 & 5.1 &9.6 & 6.0 &27.2 & 37.5 \\
            Qwen3-30B-A3B-Nex-N1 &65.3 &11.3 & 29.7 &8.3 &13.6& \textbf{51.9} \\
            InternLM3-8B & - & 0.7 & - & - & 1.6& - \\
            InternLM3-8B-Nex-N1 &63.0&8.6&20.3& 0 &0.3&44.5 \\
            \bottomrule
        \end{tabular}%
    } 
\end{table}

\paragraph{\textbf{General agentic ability.}}

We evaluate the general agentic ability of Nex-N1 on $\tau^2$-bench~\citep{barres2025tau2} and GAIA 2~\citep{andrews2025are}, following the official evaluation settings for both benchmarks. $\tau^2$-bench challenges agents with constraint-satisfaction tasks and collaboration in ``dual-control environments'', whereas GAIA 2 provides a broad assessment of end-to-end agent performance. To ensure reproducibility and stable interactions, we replace the official GPT-4o user-agent with GPT-4.1 to provide consistent checkpoint availability. Nex-N1 performs strongly on both benchmarks, showing robust general agentic abilities and effective operation in dynamic, interactive environments.

\paragraph{\textbf{Agentic coding.}}

We evaluate the agentic coding ability of Nex-N1 on realistic software engineering tasks. The SWE-bench~\citep{jimenez2024swebench} is built from real GitHub issues and their associated pull requests; the agent is given an issue description and the corresponding code repository and must generate a patch that resolves the issue, with correctness verified by unit tests. In practice, we performed experiments on the verified subset~\citep{chowdhury2024swebenchverified}. The Terminal-Bench~\citep{tbench_2025} measures the ability of an agent to perform end-to-end tasks in a terminal environment. For the above two benchmarks, we evaluate our models using an internal implementation based on the OpenHands~\citep{wang2024openhands} scaffold with maximum iterations limited to 150. We specifically examine the reliability of our model in developing real-world backend applications by evaluating its functional correctness on BaxBench~\citep{vero2025baxbench} using its official evaluation settings. The evaluation results are shown in Table~\ref{tab:main result}, where Nex-N1 shows competitive performance against existing SOTA models.

\paragraph{\textbf{Tool use.}}

We evaluate the LLM's ability to make function calls via the Berkeley Function Calling Leaderboard (BFCL) V4~\citep{patil2025bfcl}. All evaluations on Nex-N1 follow the BFCL Function Call (FC) format. We replace the search component with the Google Search API (via Serper.dev), as the community-maintained DuckDuckGo (DDG) API cannot reliably reproduce BFCL’s Web Search behavior. Using DDG led to a substantial performance drop (GPT-5 reproduced only 66.5 vs. 86.0 on the official leaderboard), while Google Search restored the expected performance (85.0), indicating that our adjustment removes discrepancies caused by the search tool. With this more stable setup, Nex-N1 exhibits competitive performance across BFCL tasks.

\subsection{Real-world Agentic Coding}

In practical scenarios, agentic coding is increasingly used for end-to-end project development, from full-stack applications to automated webpage creation. To capture these real-world demands, we design two complementary benchmarks: one measuring a model’s agentic coding ability to execute realistic project-development workflows, and another evaluating the visual fidelity and interactive quality of its web-development outputs.

\paragraph{\textbf{Project-development.}}
To evaluate the agentic coding capabilities of the model in realistic project development, we constructed a test set including 43 data samples that covered 13 distinct coding scenarios. In these test sets, we compared the Nex-N1 with the baseline model under identical configurations within the \texttt{Claude Code} framework, collecting the models' execution traces and their generated results. We evaluate their results in terms of success rate, code accuracy, execution efficiency, readability, and scenario adaptability.
The results indicate that Nex-N1 wins or ties against the major models in over half of the scenarios (e.g. $64.5\%$ against claude-sonnet-4.5, $92.9\%$ against Minimax-M2).
\begin{figure}[htbp]
    \centering
    \includegraphics[width=0.7\linewidth]{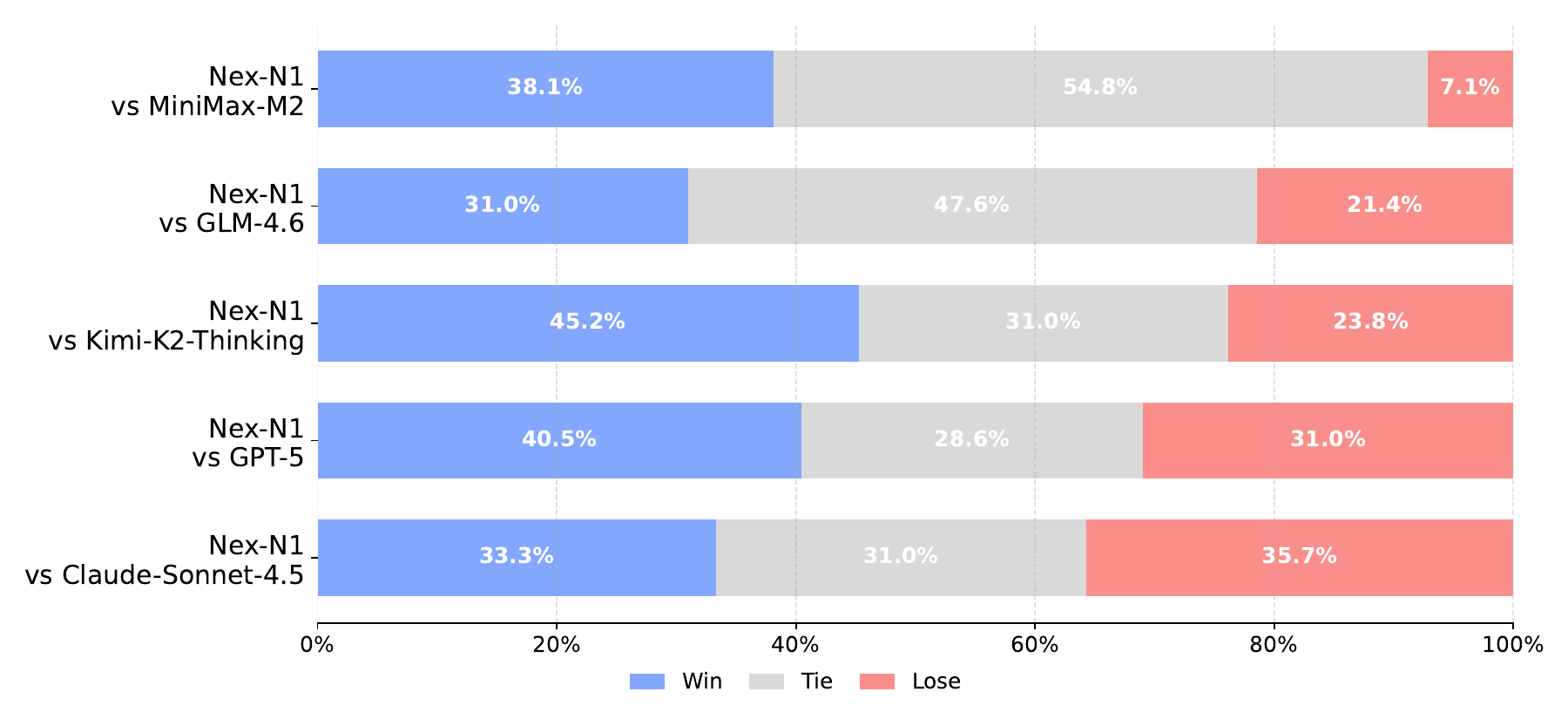}
    \caption{Human evaluation results on agentic coding.}
    \label{fig:placeholder}
\end{figure}

\paragraph{\textbf{Web-development}}
To evaluate our model's fundamental capability in developing end-to-end pages, we collected 45 data samples spanning five domains. For these samples, we compared Nex-N1 with other baseline models in a single-page format and collected the final output. We evaluate the generated pages based on visual quality, color richness, and page completeness. The results demonstrate that Nex-N1 outperforms all models except claude-sonnet-4.5 ($44.5\%$).

\begin{figure}[htbp]
    \centering
    \includegraphics[width=0.7\linewidth]{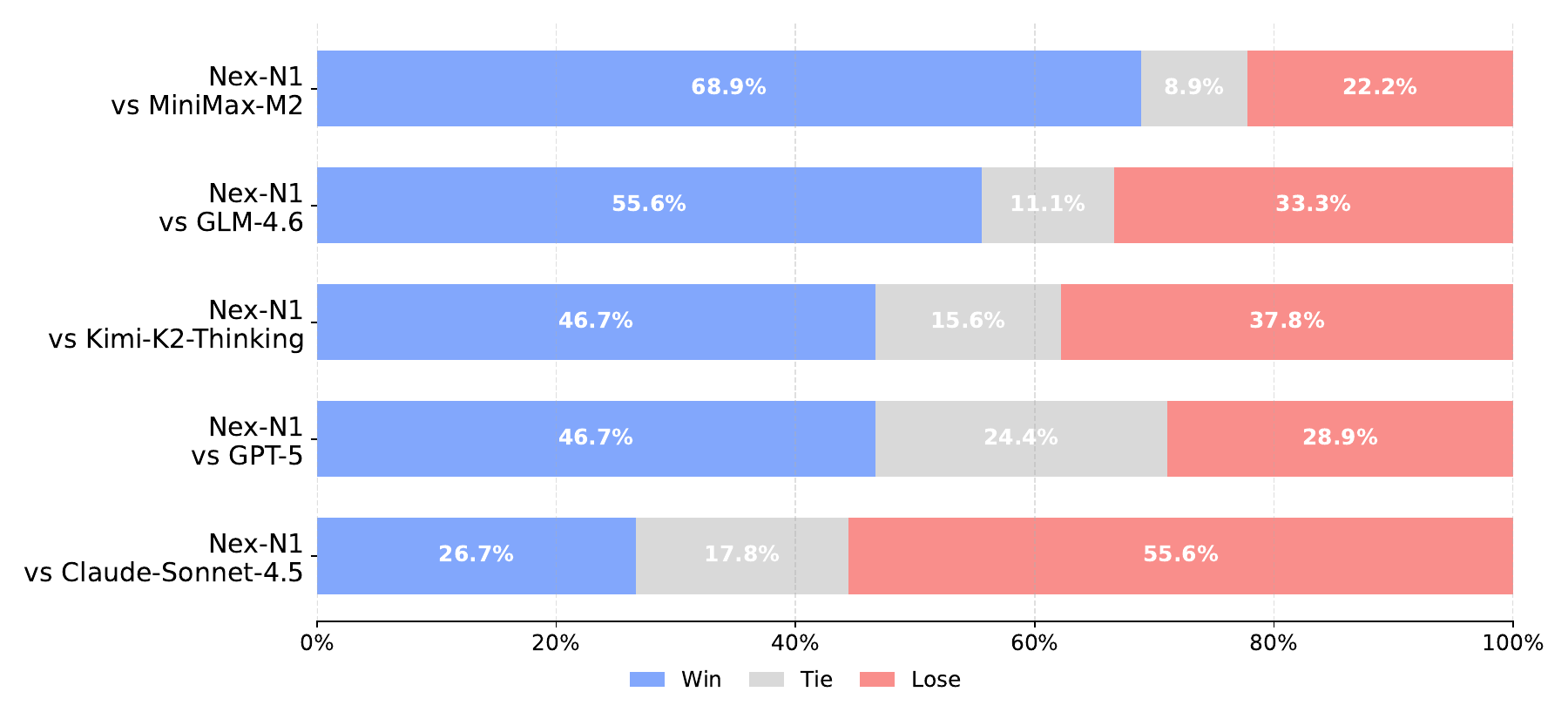}
    \caption{Human evaluation results on webpage creation.}
    \label{fig:placeholder}
\end{figure}

\subsection{Agent for Research}

Built on the NexAU framework, Nex-N1 demonstrates strong deep-research capabilities. It also agentically produces academic materials such as presentation pages and research posters.

\begin{figure}[htbp]
    \centering
    \includegraphics[width=1.0\textwidth]{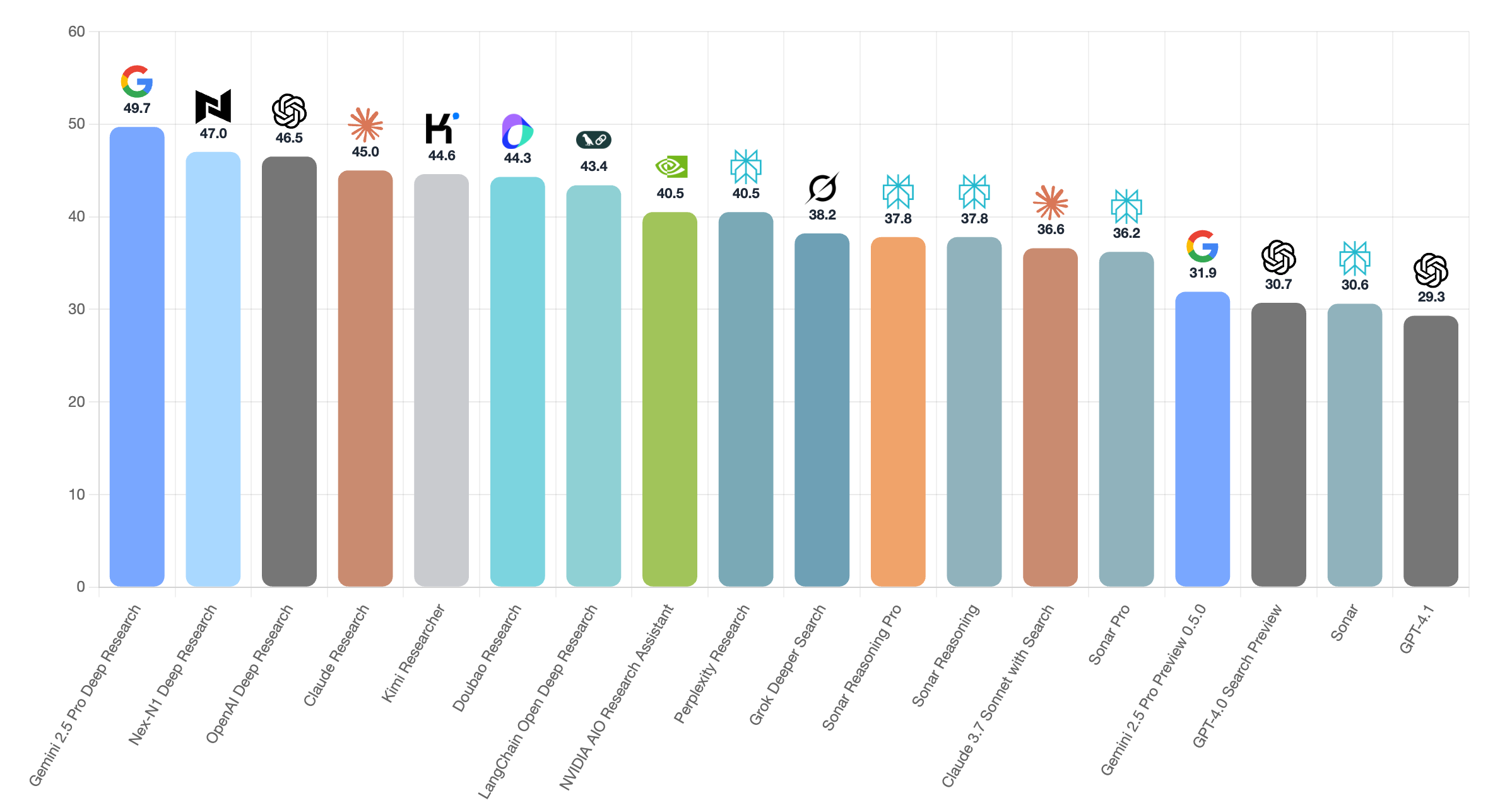}
    \caption{Result of our deep research agent on the deep research benchmark.}
    \label{fig:dr_benchmark}
\end{figure}
\paragraph{\textbf{Deep research.}}
Deep research agents have become one of the most widely used categories of AI agents, supporting information acquisition, knowledge synthesis, and automated investigative tasks across diverse workflows. To systematically evaluate our model’s real-world capabilities, we built a high-performance deep research agent\footnote{Our Deep research agent has been open-sourced at https://github.com/nex-agi/NexDR} on top of Nex-N1 and our agent framework NexAU. This system integrates Google Search, webpage visiting and parsing, and multi-step task planning, enabling the agent to autonomously execute a full research pipeline, spanning task planning, information retrieval, webpage inspection, content extraction, and iterative reflection. Through this loop, the agent can gather task-relevant information and produce high-quality, well-structured research reports for a wide range of user queries. On the public Deep Research Benchmark~\citep{du2025deepresearch}, our deep research agent built on Nex-N1 achieves a score of 47.0\%, demonstrating strong overall performance across multiple evaluation dimensions. Detailed results are provided in Fig~\ref{fig:dr_benchmark}.

\begin{figure}[htbp]
    \centering
    \includegraphics[width=1.0\textwidth]{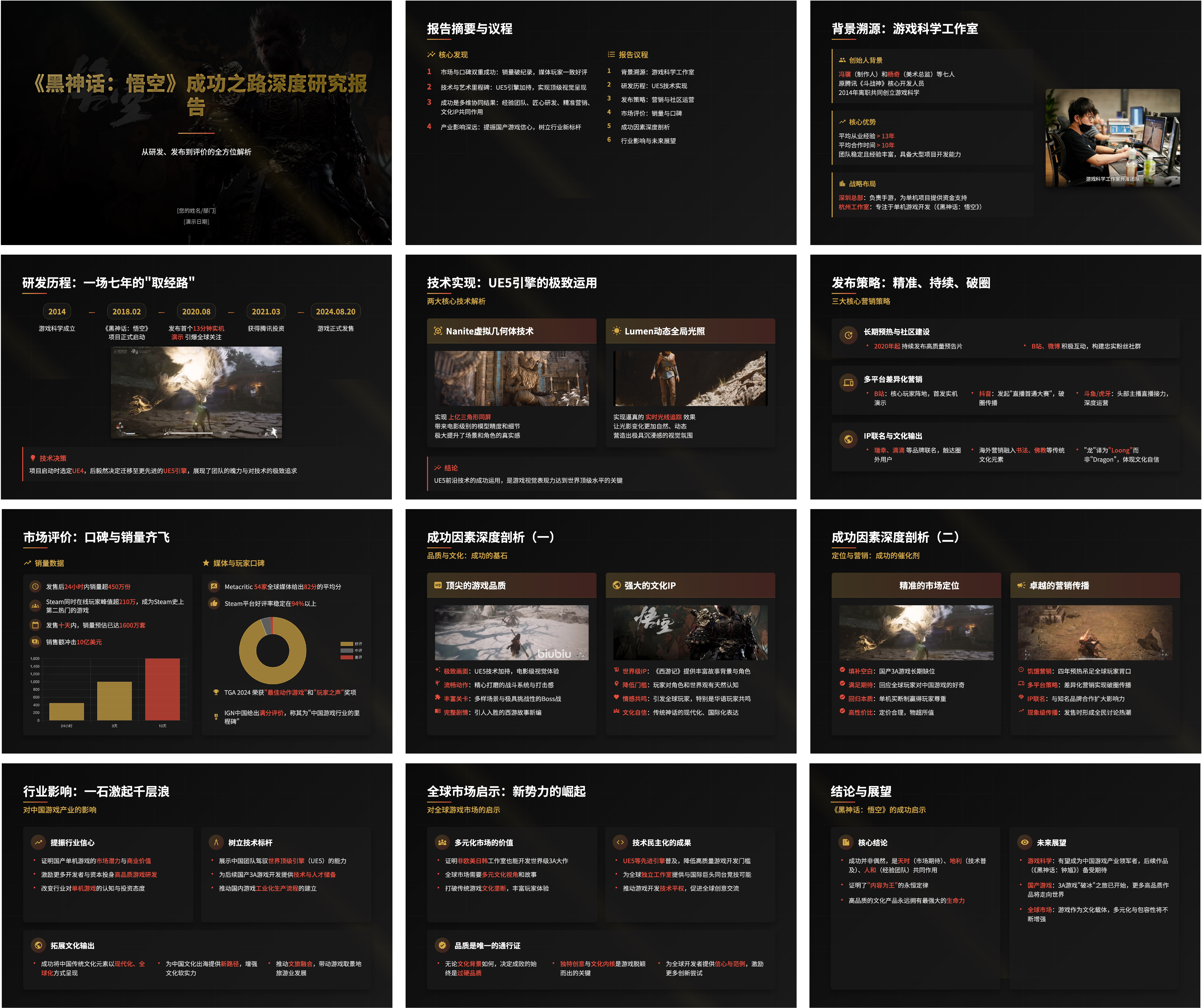}
    \caption{Demo of our deep research agent.}
    \label{fig:dr_demo}
\end{figure}

\paragraph{\textbf{Information visualization.}}
Unlike existing deep research systems such as OpenAI Deep Research or Gemini Research Agent, which output text-based reports, our deep-research agent additionally supports visualized report generation, producing both richly illustrated research reports and slide-based presentations that integrate text and graphics to highlight key findings, reasoning structure, and evidence traces. These capabilities are enabled by a dedicated sub-agent equipped with tools for image retrieval, image insertion, visual design, and slide composition. This feature significantly enhances the usability of the research outputs for communication, presentation, and downstream decision-making. A demonstration of this visualization capability is shown in Fig.~\ref{fig:dr_demo}.

\paragraph{\textbf{Poster generation.}}

Poster presentation is a vital component of academic communication. We developed the Paper2Poster Agent based on Nex-N1 and our NexAU framework, which autonomously converts papers into professional posters. The system integrates the PDF-to-Markdown parsing tool implemented with MinerU~\citep{niu2025mineru2}, the logo retrieval tool for institutions and conferences, and the QR code generation tool. Our agent is equipped with bilingual switching capability, allowing generated posters to seamlessly transition between English and Chinese versions. Additionally, we incorporated a feedback mechanism for iterative design refinement to enhance visual quality and optimize layout.

\begin{figure}[htbp]
    \centering
    \includegraphics[width=0.45\textwidth]{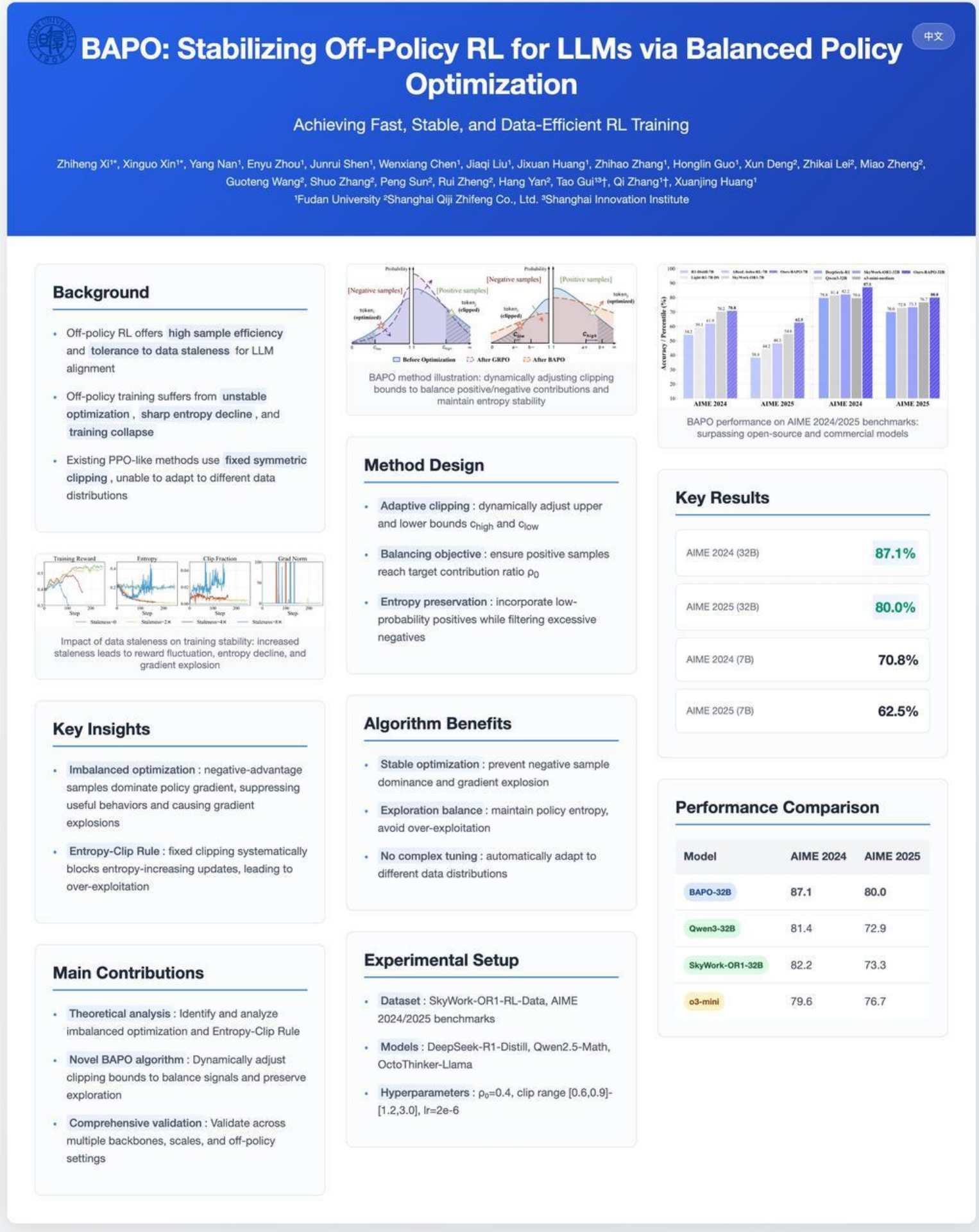}
    \hfill
    \includegraphics[width=0.52\textwidth]{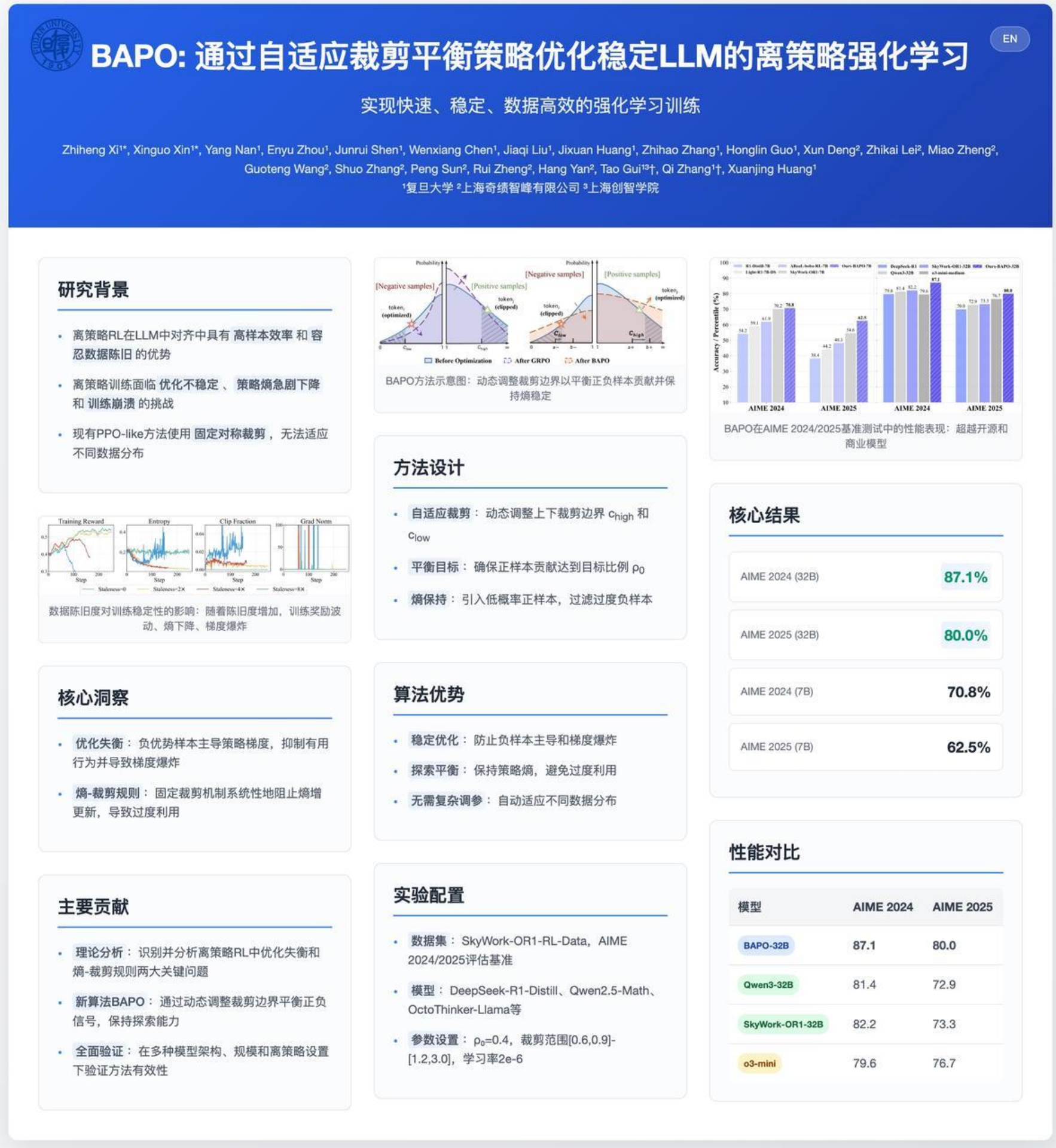}
    \caption{Demo of our Paper2Poster agent (adapted from~\citep{xi2025bapo}).}
    \label{fig:combined}
\end{figure}

\subsection{Robustness across Diverse Agent Frameworks}

We evaluate the robustness of our model across multiple agent frameworks, focusing on its atomic agentic capabilities. To this end, we measure the performance of the model in the SWE-bench~\citep{jimenez2024swebench,chowdhury2024swebenchverified} in different agent frameworks, providing a comprehensive view of its ability to generalize and execute tasks reliably in various agentic contexts. We use the Terminal Bench Harnsess~\citep{tbench_2025} for evaluation. It provides a unified interface for agentic task evaluation across multiple agent frameworks, including OpenHands~\citep{wang2024openhands}, Claude Code, and Terminus-2~\citep{tbench_2025}. Due to cost considerations, we conducted an evaluation only on a randomly sampled subset of 100 instances from the SWE-Bench verified. According to the results shown in Table~\ref{tab:robust-agent-framwork}, Nex-N1 shows a stable performance on SWE-Bench verified.~\footnote{Due to technical constraints, we only evaluated GLM-4.6, MiniMax-M2, and Nex-N1 in Claude Code. Claude Sonnet 4.5 wasn't tested on OpenHands due to sampling parameter conflicts with its API. MiniMax-M2 scored 0 on OpenHands for persistently accessing \texttt{/workspace} instead of the instructed \texttt{/testbed}.}

\begin{section}{Conclusion and Future Work}

In this work, we presented a comprehensive infrastructure for agentic scaling, enabling the systematic generation of realistic and grounded interaction trajectories. By automating the construction of agent environments through NexAU and NexA4A, we address the critical bottleneck of data scarcity, establishing a robust foundation for training agents that excel in complex tool-use scenarios.

Future work will focus on evolving this infrastructure into a large-scale simulation platform for Reinforcement Learning. We aim to automatically construct environments that are not only highly diverse and increasingly difficult but also objectively verifiable. This will allow us to move beyond static supervision, creating a dynamic ``gym'' where agents can self-improve and master long-horizon reasoning through active exploration and direct environmental feedback.

\end{section}

\begin{table}[]
    \centering
    \caption{Performance on a 100-instance subset of SWE-Bench verified across diverse agent frameworks.}
    \begin{tabular}{@{}lllll@{}}
    \toprule
                      & Terminus 2 XML & Claude Code & OpenHands & Reported \\ \midrule
    GLM 4.6           & 57.5           & 56          & 56        & 68       \\
    MiniMax M2        & 38             & 64.5        & 0         & 69.4     \\
    DeepSeek v3.1     & 43             & -           & 44.5      & 66       \\
    Kimi K2 0905      & 50.5           & -           & 48.5      & 69.2     \\
    Kimi K2 Thinking  & 62             & -           & 53.1      & 71.3     \\
    Claude Sonnet 4.5 & 57             & -           & -         & 77.2     \\
    Nex-N1            & 51.2           & 62          & 63.5      & 70.6     \\ \bottomrule
    \end{tabular}
    \label{tab:robust-agent-framwork}
\end{table}

%% file: chapters/appendix.tex
\appendix

\newpage
\section*{Project Team \& Acknowledgements}
The names are sorted in alphabetical order of the last name.
\subsection*{Contributors}
\noindent
% C
Yuxuan Cai, % 蔡於轩
Lu Chen, % 陈璐
Qiaoling Chen, % 陈巧玲
% D
Yuyang Ding, % 丁宇洋
% F
Liwen Fan, % 范力文
Wenjie Fu, % 傅文杰
% G
Yufei Gao, % 高宇菲
Honglin Guo, % 郭虹麟
Pinxue Guo, % 郭品学
% H
Zhenhua Han, % 韩震华
Zhengfu He, % 贺正夫
Hanglei Hu, % 胡航磊
Kai Hu, % 胡凯
Shengjia Hua, % 华圣佳
Tianyu Huai, % 怀天宇
Baodai Huang, % 黄宝岱
% J
Li Ji, % 纪力
Zhen Jiang, % 姜镇
% L
Zhikai Lei, % 雷智凯
Bufan Li, % 李不凡
Jiahang Lin, % 林佳航
Lizhi Lin, % 林李挚
Jinxiu Liu, % 刘锦绣
Shichun Liu, % 刘世纯
Ziming Liu, % 刘子铭
% N
Yuchen Ni, % 倪雨琛
% Q
Pengfang Qian, % 钱鹏方
% S
Yujiong Shen, % 沈钰炯
Qingyun Shi, % 石青芸
Wentao Shu, % 舒文韬
Peng Sun, % 孙鹏
Yiran Suo, % 锁祎然
% T
Tian Tang, % 唐添
Boyu Tian, % 田博宇
% W
Guoteng Wang, % 王国腾
Junzhe Wang, % 王浚哲
Peixin Wang, % 王培鑫
% X
Zhiheng Xi, % 奚志恒
% Y
Hang Yan, % 颜航
Jie Yang, % 杨捷
Zhixiong Yang, % 杨智雄
Tianchu Yao, % 姚天楚
Guangze Ye, % 叶光泽
Qianxi Yu, % 余千禧
% Z
Shuo Zhang, % 张硕
Xinyue Zhang, % 张馨月
Yiqi Zhang, % 张懿麒
Jiarong Zhao, % 赵家荣
Miao Zheng, % 郑淼
Rui Zheng, % 郑锐
Enyu Zhou, % 周恩宇
Jiazheng Zhou, % 周家正
Maosen Zhou, % 周卯森
Yuhao Zhou % 周钰皓

\subsection*{Project Lead}
Tao Gui$^\star$, 
Yining Zheng$^\star$, 
Xinchi Chen, 
Jie Zhou, 
Siyuan Feng, 
Qin Chen, 
Liang He, 
Qi Zhang, 
Xuanjing Huang, 
Xipeng Qiu

\subsection*{Affiliations}
Shanghai Innovation Institute\\
Shanghai Qiji Zhifeng Co., Ltd. \\
Fudan University \\
East China Normal University \\
MOSI Intelligence\\

\renewcommand\thefootnote{}\footnote{$^\star$Equal contribution.}

% \section*{Acknowledgments}
% Use unnumbered first level headings for the acknowledgments. All
% acknowledgments, including those to funding agencies, go at the end of the paper.

% \section*{Ethics Statement}
% Authors can add an optional ethics statement to the paper. 
% For papers that touch on ethical issues, this section will be evaluated as part of the review process. The ethics statement should come at the end of the paper. It does not count toward the page limit, but should not be more than 1 page. 